\documentclass{article}
\usepackage{spconf,amsmath,graphicx}
\usepackage{multirow}
\pdfoutput=1

\title{Encoder-decoder with Focus-mechanism for Sequence Labelling Based Spoken Language Understanding}
\name{Su Zhu and Kai Yu \thanks{This work was supported by the China NSFC project No. 61573241 and the Interdisciplinary Program (14JCZ03) of Shanghai Jiao Tong University in China.}}
\address{Key Laboratory of Shanghai Education Commission for Intelligent Interaction and Cognitive Engineering\\
    SpeechLab, Department of Computer Science and Engineering\\
    Brain Science and  Technology Research Center\\
    Shanghai Jiao Tong University, Shanghai, China\\
  {\small \tt \{paul2204,kai.yu\}@sjtu.edu.cn}
}
\begin{document}
%\ninept
%
\maketitle
\begin{abstract}
This paper investigates the framework of encoder-decoder with attention for sequence labelling based spoken language understanding. We introduce Bidirectional Long Short Term Memory - Long Short Term Memory networks (BLSTM-LSTM) as the encoder-decoder model to fully utilize the power of deep learning. In the sequence labelling task, the input and output sequences are aligned word by word, while the attention mechanism cannot provide the exact alignment. To address this limitation, we propose a novel {\em focus mechanism} for encoder-decoder framework. Experiments on the standard ATIS dataset showed that BLSTM-LSTM with focus mechanism defined the new state-of-the-art by outperforming standard BLSTM and attention based encoder-decoder. Further experiments also show that the proposed model is more robust to speech recognition errors. 
\end{abstract}
\section{Introduction}
\label{sec:intro}

In a spoken dialogue system, the Spoken Language Understanding (SLU) is a key component that parses user utterances into corresponding semantic concepts. The semantic parsing of input utterances in sequence labelling typically consists of three tasks: domain detection, intent determination and slot filling. In this paper, we focus on the sequence labelling based slot filling task which assigns a semantic slot tag for each word in the sentence. The main challenges of SLU are the performance improvement and its robustness to ASR errors.

\begin{figure}[htbp]
\centering
\includegraphics[width=8cm,height=0.9cm]{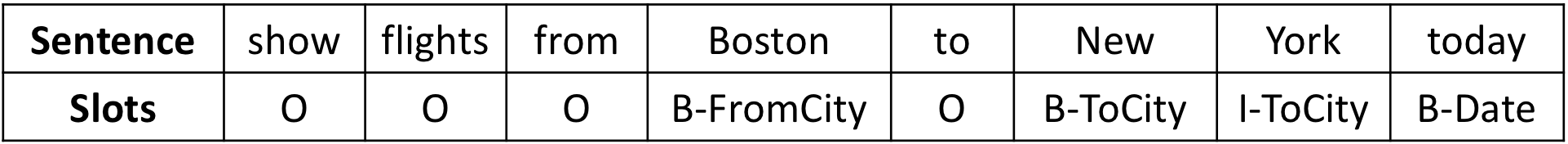}
\caption{An example of ATIS sentence and the annotated slots.}
\label{fig:atis}
\end{figure}

Slot filling is a main task of SLU to obtain semantic slots and the associated values. Typically, slot filling would be treated as a sequence labelling (SL) problem to predict the slot tag for each word in the utterance. As a typical alignment task, one example of slot filling is illustrated in Figure \ref{fig:atis}. The goal is to label the word ``\emph{Boston}" as the departure city,   ``\emph{New York}" as the arrival city, and  ``\emph{today}" as the date.

Standard approaches to solve this problem include generative models, such as HMM/CFG composite models \cite{wang2005spoken} , hidden vector state (HVS) model \cite{he2003data}, and discriminative or conditional models such as conditional random fields (CRFs) \cite{lafferty2001conditional}, and support vector machines (SVMs) \cite{taku2001chunking}. Recently, motivated by a number of very successful continuous-space, neural network and deep learning approaches \cite{mikolov2010recurrent,mikolov2013linguistic}, many neural network architectures have been applied to this task, such as simple recurrent neural networks (RNNs) \cite{yao2013recurrent,mesnil2013investigation,mesnil2015using}, convolutional neural networks (CNNs) \cite{xu2013convolutional}, long short-term memory (LSTM) \cite{yao2014spoken} and the variations of different training criterions \cite{yao2014recurrent,vu2016bi}. The most recent papers use variations on LSTM based sequence models, including encoder-decoder, external memory \cite{peng2015recurrent,kurata2016leveraging}.  %extensions \cite{peng2015recurrent,kurata2016leveraging}.%

Inspired by the success of the attention mechanism \cite{bahdanau2014neural} in Natural Language Processing (NLP) field, we first applied an attention-based encoder-decoder \cite{vinyals2015grammar} to treat the sequence labelling based SLU as a language translation problem. In order to consider the previous and the future information, we modelled the encoder with a bidirectional LSTM (BLSTM), and the decoder with an unidirectional LSTM. The attention mechanism takes the weighted average of scores provided by the matches between inputs around position $A$ and output at position $B$. There are two main limitations of attention model in sequence labelling task: 
\begin{itemize}
\item Input and output in the sequence labelling are aligned while the attention model scores the overall input words. 
\item The alignment could be learned by the attention model, but is difficult to approach with limited annotated data in sequence labelling task (unlike Machine Translation in which paired data is easier obtained). 
\end{itemize}
To address the limitations of the attention mechanism in sequence labelling, we propose the focus mechanism which is emphasizing the aligned encoder's hidden states.

The remainder of the paper is organized as follows. Section 2 discusses related research. Section 3 describes the BLSTM-LSTM based the encoder-decoder, the attention and focus mechanisms. Section 4 reports the experiment results. Finally, Section 5 draws conclusions.

\section{Related Works}
\label{sec:rw}

Recent research regarding slot filling has been focused on RNN and its extensions. At first,  \cite{yao2013recurrent} used RNN to beat CRF in the ATIS dataset.  \cite{mesnil2013investigation} tried bi-directional and hybrid  RNN to investigate using RNN for slot filling. \cite{yao2014spoken} introduced LSTM and deep LSTM architecture for this task and obtained a marginal improvement over RNN. \cite{peng2015recurrent} proposed RNN-EM which used an external memory architecture to improve the memory capability of RNN. \cite{vu2016bi} proposed to use the ranking loss function to train a bi-directional RNN. 

Except for the architectures of neural networks, many studies have been conducted to model the label dependencies. \cite{xu2013convolutional} proposed to combine CNN and CRF for sentence-level optimization.  \cite{mesnil2013investigation,liu2015recurrent} combined Elman-type and Jordan-type RNNs to consider the dependency on the last output label.

Following the success of attention based models in the NLP field,  \cite{edwin2015exploring} applied the attention-based encoder-decoder to the slot filling task, but without LSTM cells. \cite{kurata2016leveraging} proposed encoder-labeler architecture with two LSTMs which are encoder LSTM and labeler LSTM% but the encoder is not BLSTM
. The encoder-labeler model got the best performance of 95.66\% $F_1$-score in the ATIS dataset.

In order to achieve a full investigation, we combine BLSTM which considers the past and future information within the powerful encoder-decoder model to introduce the BLSTM-LSTM based encoder-decoder in sequence labelling task. % \footnote{It was put on the arXiv on Aug. 6th, 2016.}.

\section{Proposed Models}
\label{sec:pm}

%As a competitive baseline, CRF models the label relations. 
By considering the past inputs only, unidirectional LSTM cannot solve long distance dependencies of future inputs. BLSTM addressed this shortcoming with two unidirectional LSTMs: a forward pass which processes the original input word sequence; a backward pass which processes the reversed input word sequence. To learn the advantages of these models, we are going to introduce a BLSTM-LSTM based encoder-decoder architecture.
%Compared with LSTM, simple RNN is not good at keeping or forgetting history. 
%Moreover, the label dependency would be considered in our proposed model.

%\subsection{BLSTM for Slot Filling}
%\label{subsec:blstm}

%With input, output and forgetting gates, LSTM has some advanced properties compared to the simple RNN in slot filling \cite{yao2014spoken}. In uni-directional LSTM, only the past inputs are considered. Although in some previous works, they used a context window to take the next $k$ words into account, but that is limited. To regard not only the previous history but also the future history, bi-directional LSTM (BLSTM) consists of two uni-directional LSTMs: a forward passed LSTM which processes the original input word sequences; a backward passed LSTM which processes the reversed input word sequences. Then the concatenation of  the forward and backward hidden layer is fed into the output layer to predict the semantic label for the current word.

\subsection{BLSTM-LSTM + Attention}
\label{subsec:encdec}

We followed the encoder-decoder from \cite{bahdanau2014neural} which is based on RNN. To consider both the previous history and the future history, we use BLSTM as the encoder and LSTM as the decoder.

An important extension of encoder-decoder is to add an attention mechanism. We adopted the attention model from \cite{vinyals2015grammar}. The only difference is that we use BLSTM as encoder in advance. The encoder reads the input sentence $\textbf{x}=(x_1, x_2, ..., x_{T_x})$ and generates $T_x$ hidden states by BLSTM: 
\begin{equation*}
\begin{split}
h_i &= [\overleftarrow{h_{i}}, \overrightarrow{h_{i}}] \\
\overleftarrow{h_{i}} &= f_l(\overleftarrow{h_{{i+1}}}, x_i) \\
\overrightarrow{h_{i}} &= f_r(\overrightarrow{h_{{i-1}}}, x_i)
\end{split}
\end{equation*}
where $\overleftarrow{h_{i}}$ is the hidden state of backward pass in BLSTM and $\overrightarrow{h_{i}}$ is the hidden state of forward pass in BLSTM at time $i$. 

The decoder is trained to predict the next semantic label $y_t$ given the all input words and all the previously predicted semantic labels \{$y_1, ...,y_{t-1}$\} :
\begin{equation*}
\begin{split}
P(y_t|y_1, ...,y_{t-1};\textbf{x}) &= g(s_t) \\
s_t &= f_d(s_{t-1}, y_{t-1}, c_t) \\
c_t &= q(s_{t-1}, h_1, ..., h_{T_x})
\end{split}
\end{equation*}
where $g$ refers to the output layer (often with softmax) and $s_t$ is the hidden state of decoder LSTM at time $t$, with $f_d$ set as LSTM unit function. $c_t$ denotes the contextual information for generating label $y_t$ according to different encoder hidden states, which is typically implemented by an attention mechanism \cite{bahdanau2014neural}, e.g.
\begin{equation*}
\begin{split}
c_t &= \sum_{i=1}^{T_x}\alpha_{ti}h_i \\
\alpha_{ti} &= \frac{\text{exp}(a(s_{t-1}, h_i))}{\sum_{j=1}^{T_x}\text{exp}(a(s_{t-1}, h_j))}
\end{split}
\end{equation*}
where $a$ is a feed-forward neural network. $s_0$ is initialized with $\overleftarrow{h_{1}}$. In order to apply this model for sequence labelling task, we enforce the output sequence generated by the decoder to get the same length of the input word sequence.

%The architecture is depicted in Figure \ref{fig:attention}. The hidden state of decoder LSTM is initialized with the last hidden state of the backward passed LSTM in the encoder.

%\begin{figure}[htbp]
%\centering
%\includegraphics[width=5cm]{./figures/attention2.png}
%\caption{Illustration of Attention-based encoder-decoder architecture.}
%\label{fig:attention}
%\end{figure}

\subsection{Focus mechanism}
\label{subsec:focus}

\begin{figure}[htbp]
\centering
\includegraphics[width=8cm]{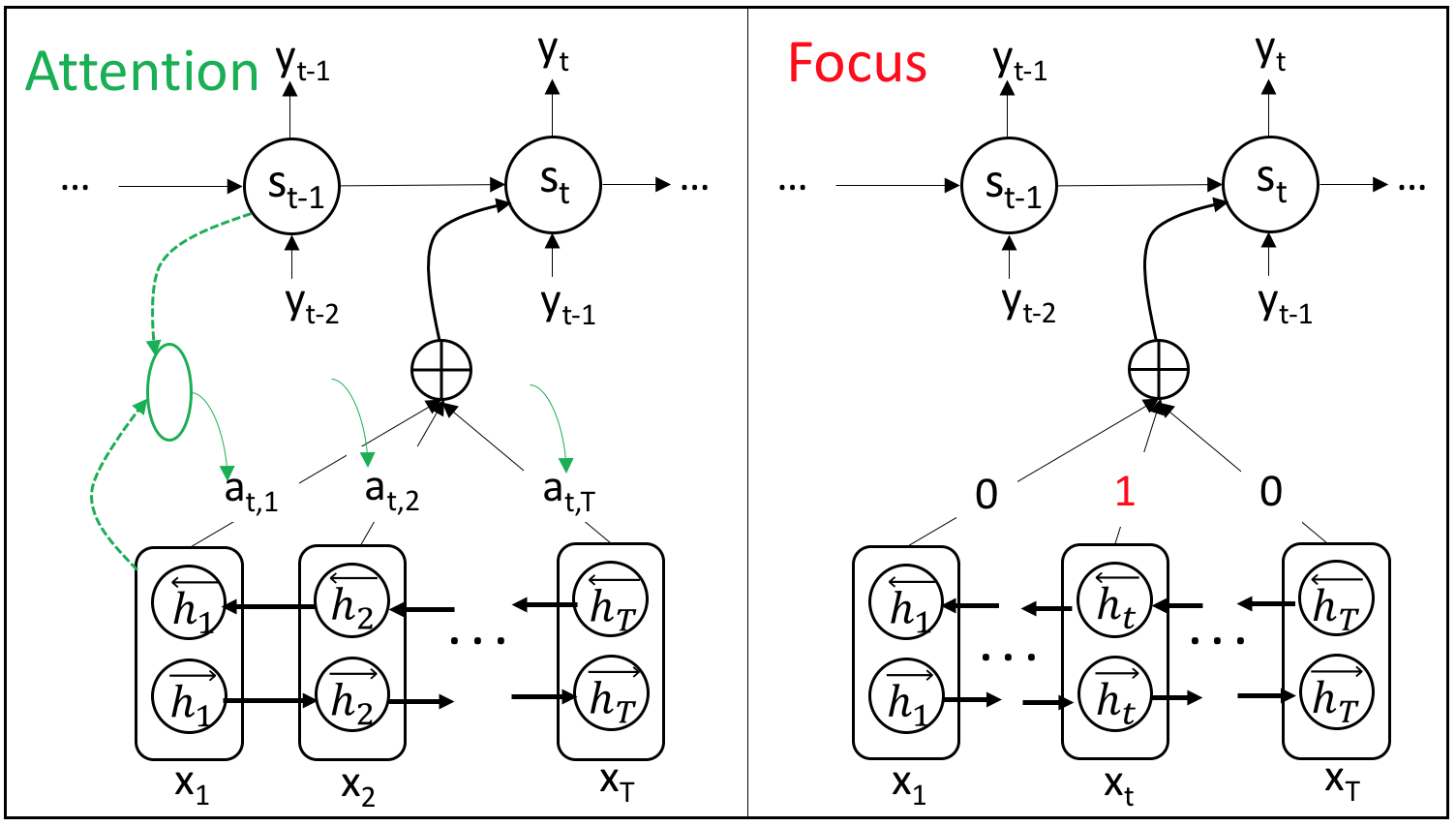}
\caption{Illustration of the attention and focus mechanism.}
\label{fig:attention_focus}
\end{figure}

As referenced in the introduction, the attention mechanism is facing with two limitations in sequence labelling based SLU task. To address these problems, we propose the focus mechanism that only considers the aligned encoder hidden state, i.e. $\alpha_{ti}=0, \text{ if }t\neq i; \alpha_{ti}=1, \text{ if }t=i$. Thus, 
\begin{equation*}
c_t = h_t
\end{equation*}
Therefore, there is no necessity to learn the alignment by utilizing the attention model. The encoder-decoder with attention and focus mechanisms are illustrated as figure \ref{fig:attention_focus}.

%Slot filling treated as a sequence labelling problem is an alignment task. Because the output and input sequences are aligned that each input word is mapped to a semantic tag (special ``O" tag for no meaning word). Thus the output and input sequences are in the same length, which is quite different from some tasks, e.g. Machine Translation, Question Answering. 
%As a prior knowledge, slot filling treated as a sequence labelling problem is an alignment task. Because the output and input sequences are aligned that each input word is mapped to a semantic tag (special ``O" tag for no meaning word). Thus the output and input sequences are in the same length. That's quite different from some tasks, e.g. Machine Translation, Question Answering. So we don't need to learn the alignment by an attention mechanism. In terms that, we simplify the attention mechanism in encoder-decoder to be a direct connection from $i$th encoder hidden vector to $i$th decoder hidden vector. To compared with attention mechanism, we call it ``focus" mechanism. This simplified seq2seq architecture is shown in Figure \ref{fig:focus}.

\section{Experiments}
\label{sec:exp}

\subsection{Experimental Setup}

We use the ATIS corpus which has been widely used as a benchmark by the SLU community. In ATIS, the sentence and its semantic slot labels are in the popular in/out/begin (IOB) representation. An example sentence is provided in figure \ref{fig:atis}. The training data consists of 4978 sentences and 56590 words. Test data consists of 893 sentences and 9198 words. We randomly selected 80\% of the training data for model training and the remaining 20\% for validation \cite{mesnil2015using}. 

In addition to ATIS, we also apply our models for a custom Chinese dataset from the car navigation domain which contains 8000 utterances for training, 2000 utterances for validation and 1944 utterances for testing. Each word has been manually assigned a slot using IOB schema. Not only the natural sentence, the top hypothesis of each utterance produced from the automatic speech recognition (ASR)  is also evaluated. These ASR top outputs have a word error rate (WER) of  4.75\% and a sentence error rate (SER) of 23.42\%.  

We report the $F_1$-score on the test set with parameters that achieved the best $F_1$-score on the validation data. We deal with unseen words in the test set by marking any words with only one single occurrence in the training set as $<unk>$.% for these two datasets. 

Our implemented LSTM neural networks are identical to the ones in  \cite{graves2013generating}. As described earlier, the encoder-decoder model utilized a BLSTM for encoding and a LSTM for decoding. For training, the network parameters are randomly initialized in accordance with the uniform distribution (-0.2, 0.2). We used the  stochastic gradient descent (SGD) for updating parameters. In order to enhance the generalization capability of our proposed models, we applied \emph{dropout} with a probability of 0.5 during the training stage. %The dropout is added onto the input and output layer.

For encoder-decoder, we used left-to-right beam searching for decoding with beam size of two empirically. We tried different learning rates, ranging from 0.004 to 0.04 similar to grid-search. We kept the learning rate for 100 epochs and saved the parameters that gave the best performance on the validation set, which is measured after each training epoch.

\subsection{Results on the ATIS Dataset}
\label{subsec:atis}

Table \ref{tab:results} shows the results on ATIS dataset. For all architectures, we set the dimension of word embeddings to 100 and the number of hidden units to 100. We only use the current word as input without any context words. BLSTM, which considers both the past and the future history, outperforms LSTM (+2.03\%). The \emph{attention based BLSTM-LSTM} model got lower F1-score than \emph{BLSTM} (-2.7\%). We think  the reason is that the sequence labelling problem is a task, whose input and output sequences are aligned. 

Having only limited data, it is difficult to learn the alignment accurately by using the attention mechanism. We try to expand the training data of ATIS by randomly replacing the value of each specific slot within sentences to 10 times that of the original scale. For example, ``Flights from Boston" can be expanded to ``Flights from New York", ``Flights from Los Angeles", etc. The BLSTM-LSTM with attention achieves a 95.19\% $F_1$-score, while other methods did not benefit from the expanded training set.

\begin{table} [htbp!]
\vspace{2mm}
\centerline{
\small
\begin{tabular}{c|c||c}
\hline
Model & Mechanism &  $F_1$-score (\%) \\
\hline  \hline
%CRF \cite{mesnil2013investigation} & 92.94   \\
%\hline
LSTM & - & 93.40 \\
%\hline
BLSTM & - & 95.43   \\
\hline
\multirow{2}{*}{BLSTM-LSTM} & \tt Attention & 92.73 \\
 \cline{2-3}
 & \tt Focus & \textbf{95.79} \\
\hline
\end{tabular}
}
\caption{\label{tab:results} {\it Experimental results on ATIS dataset.}}
\end{table}

By considering the alignment of the sequence labelling task, the \emph{BLSTM-LSTM with focus} increased the F1-score from 92.73\% to 95.79\% and achieved an 0.36\% improvement (significant level 10\%) in comparison to \emph{BLSTM}.  We think the \emph{BLSTM-LSTM with focus} has two advantages over the \emph{BLSTM}: 1) the initialization of hidden state of decoder LSTM with $s_0=\overleftarrow{h_{1}}$ provides sentence leveraging features; 2) it enables label dependency within the decoder.

%1) using the last hidden state of backward passed LSTM in encoder as a sentence leveraging features for the initilization of hidden state of decoder LSTM ; 2) label dependency in the decoder.  

%When any of these two advantages is omitted, the performance get decreased (except \emph{Encder-decoder (focus), no label dep.} in Table \ref{tab:results} doesn't change the $F_1$-score).

%To analysis the improvement, we closed the initialization link from the left of encoder to the right of decoder (\emph{no link}) as shown in \ref{fig:focus} and took the label dependency away from the decoder layer (\emph{no label dep.}). It seems the link helps a lot. Following \cite{bahdanau2014neural}, we add a linear transformation and \emph{tanh} activation onto the initialization link from encoder to decoder. The result is shown as the last line of Table \ref{tab:results}.
%It seems the link helps a lot. Finally, we tried the combination of aligned hidden states and context feature from attention model, it didn't take any improvement compared with \emph{Bi-LSTM+Decoder layer (aligned)}.

\begin{table} [htbp!]
\vspace{2mm}
\centerline{
\small
\begin{tabular}{lc}
\hline
Model & $F_1$-score \\
\hline  \hline
CRF \cite{mesnil2013investigation} & 92.94   \\
simple RNN \cite{yao2013recurrent} & 94.11   \\
CNN-CRF \cite{xu2013convolutional} & 94.35   \\
LSTM \cite{yao2014spoken} & 94.85   \\
RNN-SOP \cite{liu2015recurrent} & 94.89   \\
Deep LSTM \cite{yao2014spoken} & 95.08   \\
RNN-EM \cite{peng2015recurrent} & 95.25   \\
Bi-RNN with Ranking Loss \cite{vu2016bi} & 95.47 \\
 Encoder-labeler Deep LSTM \cite{kurata2016leveraging}  & 95.66 \\
\hline
\textbf{BLSTM-LSTM (focus)}  & \textbf{95.79}    \\
%\textbf{BLSTM-LSTM (focus) + Ranking Loss}  & \textbf{x}    \\
\hline
\end{tabular}
}
\caption{\label{tab:compare} {\it Comparison with published results on ATIS.}}
\end{table}

%To seek for better performance of our methods, we further conducted hyper-parameter search for \emph{BLSTM-LSTM with focus} model. We tuned the dimension of word embedding in \{50, 75, 100\}, dimension of hidden states in \{100, 150, 200, 250, 300\}, size of context window from only the current word to additional 2 previous and 2 next words. 

Compared with the published results on the ATIS dataset, our method outperforms the previously published F1-score, illustrated in Table \ref{tab:compare}. Table \ref{tab:compare} summarizes the recently published results on the ATIS slot filling task and compares them with the results of our proposed methods. Our proposed model achieves state-of-the-art performance \footnote{There are other published results that achieved better performance by using Name Entity features, e.g. \cite{mesnil2013investigation} achieved 96.24\% $F_1$-score. The NE features are annotated and really strong. If only using NE features, BLSTM obtained 97.00\% $F_1$-score. So it would be more meaningful to use only lexicon features.} but not statistically significant. %, especially with the ranking loss criterion \cite{vu2016bi}.

%The encoder-labeler method \cite{kurata2016leveraging}  is similar to our method with encoder and decoder LSTMs also. One difference is that we use Bi-LSTM for encoder and the hidden states in encoder are fed into the decoder layer in alignment with the position, but in encoder-labeler only the last hidden states are used for the initialization of the labeler and the words are fed into the labeler directly. Another difference is the optimization method used that they used the newest ADAM \cite{kingma2014adam} but we used the standard SGD. At next step, we would like to try ADAM optimizer.

\subsection{Results on Chinese Navigation Dataset}

To investigate the robustness of the BLSTM-LSTM architectures with the attention or focus mechanism, we conduct additional experiments on the Chinese navigation dataset described in the experimental setup. For the neural network architectures, we also set the dimension of word embeddings to 100 and the number of hidden units to 100. Additionally, only the current word is used as LSTM input, in comparison to CRF which used a context window size of 5. We train the model on natural text sentences (without any speech recognition errors) and test it on not only manual transcriptions (correct text sentences), but also top hypotheses from speech recognition systems (including recognition errors).  

\begin{table} [htbp!]
\vspace{2mm}
\centerline{
\small
\begin{tabular}{c|c||c|c}
\hline
Model & Mechanism &  Manual Trans.  & ASR Hyp.\\
\hline  \hline
CRF & - & 94.55 & 91.51  \\
%\hline
LSTM & - & 79.90 & 74.25 \\
%\hline
BLSTM & - & 95.33 & 91.23  \\
\hline
\multirow{2}{*}{BLSTM-LSTM} & \tt Attention &  95.65 & 91.76 \\
 \cline{2-4}
 & \tt Focus & \textbf{96.60} & \textbf{93.08} \\
%Encder-decoder (attention) & 95.07 & 91.07   \\
%\hline
%Encder-decoder (focus) & \textbf{96.26} & \textbf{92.54}  \\
%\& no link & 95.06 & 92.48  \\
%\& no label dep. & 95.56 & 91.76   \\
%\hline
%+ link transform & \textbf{96.52}  & \textbf{93.22}  \\
\hline
\end{tabular}
}
\caption{\label{tab:navi} {\it $F_1$-scores of manual transcription and top hypothesis from ASR  on Navigation dataset.}}
\end{table}
Table \ref{tab:navi} shows the results. CRF baseline seems competitive to BLSTM, due to the sentence-level optimization of the output. In comparison, the LSTM does not meet our expectations. Because the main challenge in this dataset is detecting longer phrases like location name (the length varies from 1 to 24 words). It suffers from long distant dependencies on past and future inputs. Subsequently, BLSTM solves this problem.

BLSTM-LSTM with focus-mechanism outperforms BLSTM on both natural sentences and top hypotheses from ASR significantly (significant level 5\%). It seems BLSTM-LSTM encoder-decoder with focus mechanism is more robust to ASR errors. A possible reason is, that the label dependency in the decoder helps omit the error transformed from the encoder. CRF also models label dependency and outperforms BLSTM by parsing ASR outputs.

%\subsection{Analysis}

%In the terms of performances on test set, (\emph{Encder-decoder (focus) + link transform}) adding a transformation referenced in Section \ref{subsubsec:atis} seems to get better $F_1$-score. We think the parameter set of encoder and decoder LSTM are different, so a transformation for the recurrent link is meaningful.

\section{Conclusions}
\label{sec:cc}

In our study, we have applied multiple BLSTM-LSTM encoder-decoders with attention and focus mechanism to SLU slot filling task. The BLSTM-LSTM architecture with focus mechanism achieved a state-of-the-art result on the ATIS dataset and shows to be robust to the ASR errors on a custom dataset. We also revealed that the attention mechanism needs more data to learn the alignment, while the focus mechanism has considered the alignment property of the sequence labelling problem. In future, we want to investigate BLSTM-LSTM with focus mechanism to other sequence labelling tasks (e.g. part-of-speech tagging, named entity recognition). Furthermore, we plan to use attention based BLSTM-LSTM for solving the SLU task in cases data is only provided unaligned.

\bibliographystyle{IEEEbib}
\bibliography{refs}

\begin{thebibliography}{10}

\bibitem{wang2005spoken}
Ye-Yi Wang, Li~Deng, and Alex Acero,
\newblock ``Spoken language understanding,''
\newblock {\em Signal Processing Magazine, IEEE}, vol. 22, no. 5, pp. 16--31,
  2005.

\bibitem{he2003data}
Yulan He and Steve Young,
\newblock ``A data-driven spoken language understanding system,''
\newblock in {\em IEEE Workshop on Automatic Speech Recognition and
  Understanding.} IEEE, 2003, pp. 583--588.

\bibitem{lafferty2001conditional}
John Lafferty, Andrew McCallum, and Fernando~CN Pereira,
\newblock ``Conditional random fields: Probabilistic models for segmenting and
  labeling sequence data,''
\newblock in {\em ICML}, 2001.

\bibitem{taku2001chunking}
K~Taku and M~Yuji,
\newblock ``Chunking with support vector machine,''
\newblock in {\em Proceedings of North American chapter of the association for
  computational linguistics}, 2001, pp. 192--199.

\bibitem{mikolov2010recurrent}
Tomas Mikolov, Martin Karafi{\'a}t, Lukas Burget, Jan Cernock{\`y}, and Sanjeev
  Khudanpur,
\newblock ``Recurrent neural network based language model.,''
\newblock in {\em INTERSPEECH}, 2010, vol.~2, p.~3.

\bibitem{mikolov2013linguistic}
Tomas Mikolov, Wen-tau Yih, and Geoffrey Zweig,
\newblock ``Linguistic regularities in continuous space word
  representations.,''
\newblock in {\em HLT-NAACL}, 2013, pp. 746--751.

\bibitem{yao2013recurrent}
Kaisheng Yao, Geoffrey Zweig, Mei-Yuh Hwang, Yangyang Shi, and Dong Yu,
\newblock ``Recurrent neural networks for language understanding.,''
\newblock in {\em INTERSPEECH}, 2013, pp. 2524--2528.

\bibitem{mesnil2013investigation}
Gr{\'e}goire Mesnil, Xiaodong He, Li~Deng, and Yoshua Bengio,
\newblock ``Investigation of recurrent-neural-network architectures and
  learning methods for spoken language understanding.,''
\newblock in {\em INTERSPEECH}, 2013, pp. 3771--3775.

\bibitem{mesnil2015using}
Gr{\'e}goire Mesnil, Yann Dauphin, Kaisheng Yao, Yoshua Bengio, Li~Deng, Dilek
  Hakkani-Tur, Xiaodong He, Larry Heck, Gokhan Tur, Dong Yu, et~al.,
\newblock ``Using recurrent neural networks for slot filling in spoken language
  understanding,''
\newblock {\em IEEE/ACM Transactions on Audio, Speech, and Language
  Processing}, vol. 23, no. 3, pp. 530--539, 2015.

\bibitem{xu2013convolutional}
Puyang Xu and Ruhi Sarikaya,
\newblock ``Convolutional neural network based triangular crf for joint intent
  detection and slot filling,''
\newblock in {\em 2013 IEEE Workshop on Automatic Speech Recognition and
  Understanding (ASRU)}. IEEE, 2013, pp. 78--83.

\bibitem{yao2014spoken}
Kaisheng Yao, Baolin Peng, Yu~Zhang, Dong Yu, Geoffrey Zweig, and Yangyang Shi,
\newblock ``Spoken language understanding using long short-term memory neural
  networks,''
\newblock in {\em 2014 IEEE Spoken Language Technology Workshop (SLT)}. IEEE,
  2014, pp. 189--194.

\bibitem{yao2014recurrent}
Kaisheng Yao, Baolin Peng, Geoffrey Zweig, Dong Yu, Xiaolong Li, and Feng Gao,
\newblock ``Recurrent conditional random field for language understanding,''
\newblock in {\em 2014 IEEE International Conference on Acoustics, Speech and
  Signal Processing (ICASSP)}. IEEE, 2014, pp. 4077--4081.

\bibitem{vu2016bi}
Ngoc~Thang Vu, Pankaj Gupta, Heike Adel, and Hinrich Sch{\"u}tze,
\newblock ``Bi-directional recurrent neural network with ranking loss for
  spoken language understanding,''
\newblock in {\em 2016 IEEE International Conference on Acoustics, Speech and
  Signal Processing (ICASSP)}. IEEE, 2016.

\bibitem{peng2015recurrent}
Baolin Peng, Kaisheng Yao, Li~Jing, and Kam-Fai Wong,
\newblock ``Recurrent neural networks with external memory for spoken language
  understanding,''
\newblock in {\em Natural Language Processing and Chinese Computing}, pp.
  25--35. Springer, 2015.

\bibitem{kurata2016leveraging}
Gakuto Kurata, Bing Xiang, Bowen Zhou, and Mo~Yu,
\newblock ``Leveraging sentence-level information with encoder lstm for natural
  language understanding,''
\newblock {\em arXiv preprint arXiv:1601.01530}, 2016.

\bibitem{bahdanau2014neural}
Dzmitry Bahdanau, Kyunghyun Cho, and Yoshua Bengio,
\newblock ``Neural machine translation by jointly learning to align and
  translate,''
\newblock {\em arXiv preprint arXiv:1409.0473}, 2014.

\bibitem{vinyals2015grammar}
Oriol Vinyals, {\L}ukasz Kaiser, Terry Koo, Slav Petrov, Ilya Sutskever, and
  Geoffrey Hinton,
\newblock ``Grammar as a foreign language,''
\newblock in {\em Advances in Neural Information Processing Systems}, 2015, pp.
  2755--2763.

\bibitem{liu2015recurrent}
Bing Liu and Ian Lane,
\newblock ``Recurrent neural network structured output prediction for spoken
  language understanding,''
\newblock in {\em Proc. NIPS Workshop on Machine Learning for Spoken Language
  Understanding and Interactions}, 2015.

\bibitem{edwin2015exploring}
Edwin Simonnet, Nathalie Camelin, Paul Deléglise, and Yannick Estève,
\newblock ``Exploring the use of attention-based recurrent neural networks for
  spoken language understanding,''
\newblock in {\em Machine Learning for Spoken Language Understanding and
  Interaction NIPS 2015 workshop (SLUNIPS 2015)}, Montreal (Canada), 11 dec.
  2015.

\bibitem{graves2013generating}
Alex Graves,
\newblock ``Generating sequences with recurrent neural networks,''
\newblock {\em arXiv preprint arXiv:1308.0850}, 2013.

\end{thebibliography}

\end{document}